\documentclass[runningheads]{llncs}

\usepackage{eccv}

\usepackage{eccvabbrv}

\usepackage{booktabs}

\usepackage[accsupp]{axessibility}  

\usepackage[numbers]{natbib}
\usepackage{graphicx}
\usepackage{wrapfig}
\usepackage{algorithm}
\usepackage{algorithmic}
\usepackage[algo2e]{algorithm2e}
\usepackage{csquotes}

\usepackage[utf8]{inputenc} %
\usepackage[T1]{fontenc}    %
\usepackage{url}            %
\usepackage{booktabs}       %
\usepackage{amsfonts}       %
\usepackage{nicefrac}       %
\usepackage{microtype}      %
\usepackage{float}         %
\usepackage{multirow}      %

\usepackage{fontawesome5}
\newcommand{\usym}[1]{\resizebox{0.5em}{!}{\faCheck}}

\SetAlFnt{\tiny}

\newcommand{\secref}[1]{\S\ref{#1}} 
\newcommand{\figref}[1]{Fig.~\ref{#1}}
\newcommand{\tabref}[1]{Tab.~\ref{#1}}

\newcommand{\objref}[1]{Obj.~\ref{#1}}

\newcommand{\proposedfed}{SuperFedNAS\xspace}

\newcommand{\spatioTempHeuristic}{MaxNet\xspace}
\newcommand{\proposedTraining}{MaxNet\xspace}
\newenvironment{tightitemize}{%
\begin{list}{$\bullet$}{%
\setlength{\itemsep}{1.5pt}%
\setlength{\topsep}{2pt}%
\setlength{\parskip}{0pt}%
\setlength{\parsep}{0pt}%

\setlength{\labelwidth}{0pt}%
\setlength{\leftmargin}{4pt}%
\setlength{\labelsep}{0pt}%
\setlength{\listparindent}{0pt}%
}}%
{\end{list}}

\newcommand{\myparagraph}[1]{\noindent{\bfseries #1.}}

\usepackage[pagebackref,breaklinks,colorlinks]{hyperref}

\usepackage{orcidlink}

\begin{document}

\definecolor{codegreen}{rgb}{0,0.6,0}
\definecolor{codegray}{rgb}{0.5,0.5,0.5}
\definecolor{codepurple}{rgb}{0.58,0,0.82}
\definecolor{backcolour}{rgb}{0.95,0.95,0.92}
\definecolor{codeindigo}{RGB}{75,0,130}
\definecolor{coderoyalblue}{RGB}{65,105,225}
\definecolor{codebrown}{rgb}{0.6,0.6,0}

 \renewcommand{\algorithmiccomment}[1]{\footnotesize\textcolor{blue}{// #1}}

\newif\ifcommenton
 \commentonfalse   
\ifcommenton

\newcommand{\alind}[1]{\textcolor{codegreen}{[Alind: #1]}}
\newcommand{\alexey}[1]{\textcolor{codeindigo}{[AT: #1]}}
\newcommand{\animesh}[1]{\textcolor{codepurple}{[AN: #1]}}
\newcommand{\hugo}[1]{\textcolor{coderoyalblue}{[Hugo: #1]}}
\newcommand{\payman}[1]{\textcolor{codebrown}{[Payman: #1]}}
\newcommand{\myungjin}[1]{\textcolor{codebrown}{[Myungjin: #1]}}
\else
\newcommand{\alexey}[1]{}
\newcommand{\alind}[1]{}
\newcommand{\animesh}[1]{}
\newcommand{\hugo}[1]{}
\newcommand{\payman}[1]{}
\newcommand{\myungjin}[1]{}
\fi

\title{\proposedfed: Cost-Efficient Federated Neural Architecture Search  for On-Device Inference} 

    \titlerunning{\proposedfed: Cost-Efficient Federated NAS for On-Device Inference}

\author{Alind Khare\inst{1} \and
Animesh Agrawal\inst{1} \and
Aditya Annavajjala \inst{1} \and 
Payman Behnam \inst{1} \and 
Myungjin Lee \inst{2} \and 
Hugo Latapie \inst{2} \and 
Alexey Tumanov \inst{1}}

\authorrunning{A. Khare et al.}

\institute{Georgia Institute of Technology, Atlanta, USA
 \and
Cisco Research, USA}

\maketitle

\begin{abstract}
Neural Architecture Search (NAS) for Federated Learning (FL) is an emerging field. It automates the design and training of Deep Neural Networks (DNNs) when data cannot be centralized due to privacy, communication costs, or regulatory
restrictions. Recent federated NAS methods not only reduce manual effort but also help achieve higher accuracy than traditional FL methods like FedAvg.
Despite the success, 
existing federated NAS methods still fall short in satisfying diverse deployment targets common in on-device inference like hardware, latency budgets, or variable battery levels. Most federated NAS methods search for 
only a limited range of neuro-architectural patterns, 
repeat them in a DNN, thereby restricting achievable performance. 
Moreover, these methods incur prohibitive training costs to satisfy deployment targets. They perform the training and search of DNN architectures repeatedly for each case. 
\proposedfed addresses these challenges by decoupling the training and search in federated NAS. \proposedfed co-trains a large number of diverse DNN architectures contained inside one supernet in the FL setting. Post-training, clients perform NAS locally to find specialized DNNs by extracting different parts of the trained supernet with no additional training. \proposedfed takes $O(1)$ (instead of $O(N)$) cost to find specialized DNN architectures in FL for any $N$ deployment targets. As part of \proposedfed, we introduce \proposedTraining---a novel FL training algorithm that performs multi-objective federated optimization of a large number of DNN architectures ($\approx 5*10^8$) under different client data distributions. Overall, \proposedfed achieves upto $37.7$\% higher accuracy for the same MACs or upto $8.13$x reduction in MACs for the same accuracy than existing federated NAS methods.

\end{abstract}

\section{Introduction}
\label{sec:intro}
Federated Learning (FL) is increasingly used in numerous applications   \cite{gboard, siri, fl_pharma, medical-imaging, fed_ehr1, fl_recommending}. 
In FL, a large number of clients collaboratively participate in a distributed training of a deep neural network (DNN) while keeping their data private \cite{fedavg, fed_distill, fedma, scaffold, fed_strategy}. FL offers three key benefits:
a) smaller communication costs,
b) massive parallelism, and
c) privacy preservation. 
Despite achieving notable success, the majority of FL works \cite{fedavg, feddyn, fedma, fedprox} rely on manually designed predefined DNN architectures, a practice that can often be sub-optimal. These predefined architectures often struggle to adapt to the nuances present in diverse data distributions across clients \cite{fednas}, leading to sub-optimal accuracy. When these DNN architectures get optimal accuracy, it comes at the expense of increased model complexity \cite{fbnet} as they are primarily designed to increase accuracy. This makes manually designed DNNs unfit for clients' on-device inference: they don't provide optimal accuracy under different deployment targets such as battery conditions, hardware, latency/MACs, memory constraints prevalent in on-device inference  \cite{nestdnn, dynamic_contrains_smartphone, slimmable, slimmable_improved}. To perform efficient inference, there is a need to automate the design and training of DNN architectures in FL.


Recent FL works partially address these limitations by proposing neural architecture search (NAS) methods \cite{fednas, fedpnas}. These methods
automatically find the most accurate DNN. 
By adapting the DNN architecture to different clients' data distributions, these methods improve accuracy over traditional FL methods like FedAvg \cite{fedavg}. 
However, when it comes to providing optimal DNN architectures for efficient inference, these methods face the following challenges:
\begin{displayquote}
\setlength{\parskip}{0pt}%
\setlength{\parsep}{0pt}%
\vspace{-0.2cm}
(C1) Existing federated NAS methods are prohibitively expensive to satisfy multiple deployment targets in on-device inference.
\vspace{-0.2cm}
\end{displayquote}

\begin{wrapfigure}{r}{0.5\columnwidth}
\vspace{-0.9cm}
 \centering
   \includegraphics[width=0.5\columnwidth]{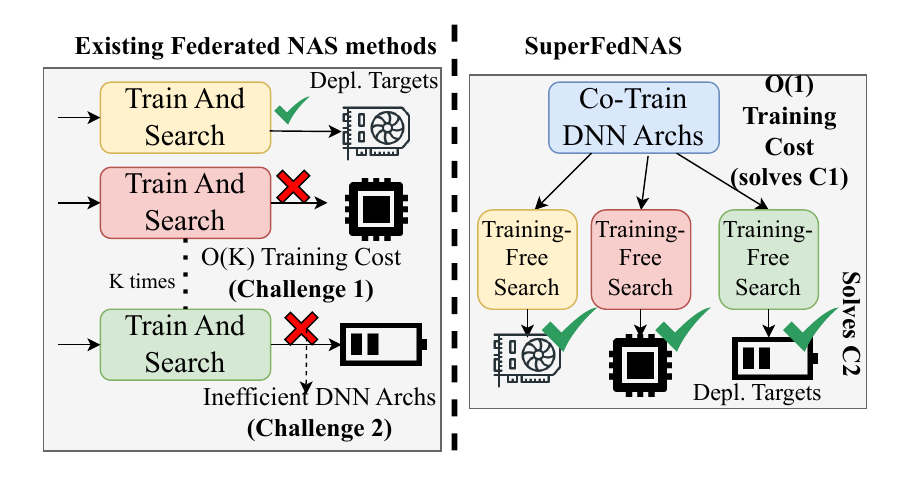}
   \caption{\proposedfed vs Existing Federated NAS methods \cite{fednas, fedpnas}}
   \label{fig:overll_dataflow}
   \vspace{-0.85cm}
\end{wrapfigure}

\noindent The existing federated NAS methods \cite{ fednas, fedpnas} only satisfy one deployment target at a time (\figref{fig:overll_dataflow}). 
These methods perform the search and training procedures simultaneously \cite{fednas}.
In the end,  the output is a single best-performing architecture subject to a deployment target. 
Therefore, for \textit{N}  
deployment targets, they need to repeat the entire process of search and training \textit{N times}. Hence, their communication and computational cost in training becomes \textit{O(N)}. These methods are simply not scalable to satisfy increasingly diverse deployment targets. 
This motivates the need to make federated NAS methods \textit{scalable}. 
Indeed, diverse deployment targets at inference are increasingly common  \cite{nestdnn, dynamic_contrains_smartphone, slimmable, slimmable_improved}. 
For instance, the GBoard application \cite{gboard} that uses the FL-trained DNNs for next-word prediction runs on a range of hardware from Apple's M-series chips with a dedicated neural engine \cite{apple_m2} to old hardware like Pixel XL (gen1) \cite{pixel_xl_gen1}. Even on the same hardware, variable resource availability (e.g., battery levels) at inference makes optimal DNN architectures significantly different.
\begin{displayquote}
\vspace{-0.22cm}
(C2) Existing federated NAS methods struggle to produce optimal DNN architectures under inference deployment targets.
\vspace{-0.22cm}
\end{displayquote}
Federated NAS methods like FedNAS \cite{fednas} tend to sub-optimally increase latency/MACs or memory to achieve better accuracy. 
Precisely, these methods only search for a building block in the DNN architecture and then repeatedly stack the most performing block to create the final DNN. Repeated stacking not only restricts the block diversity but also sub-optimally inflates latency or memory. Instead, to find optimal DNNs under deployment constraints, the federated NAS methods need to support rich diversity in DNN architectures, including diversity in blocks.
This implies that federated NAS methods need to search for optimal DNNs from a larger search space of DNN architectures.  

To solve the aforementioned challenges (\textbf{C1}, \textbf{C2}), we propose \proposedfed : a federated method that performs NAS over a rich diversity of DNN architectures for efficient on-device inference (\figref{fig:overll_dataflow}). Taking inspiration from centralized NAS techniques \cite{ofa}, \proposedfed addresses  \textbf{C1} by decoupling the training and search of DNN architectures in FL. \proposedfed has one FL training stage that co-trains a large number of DNNs simultaneously. 
It achieves this by performing FL-training of the supernet \cite{ofa}: a DNN that contains multiple smaller DNNs (subnets) with varied shapes and sizes within it. 
These subnets partially share their weights. 
Post the FL training stage, a client performs NAS locally without any additional training, and thereby has negligible cost. 
Given a deployment target, local NAS finds specialized DNNs (subnets) using predictor-guided search \cite{predictor_guided_search}. Overall, \proposedfed reduces the cost to find specialized DNNs in FL for $N$ deployment targets to $O(1)$. \proposedfed solves  \textbf{C2} as its DNN architectures support different depths and block diversity\footnote{Each block allowed to have different width.}. In fact, \proposedfed performs federated NAS over $\approx 5*10^8$ diverse DNN architectures.

However, decoupling the training and search of DNN architectures in federated NAS is non-trivial. It involves federated co-training of a large number of subnets in the supernet across multiple clients. Such training needs to perform multi-objective federated optimization of the shared weights in the supernet and optimize the accuracy of each subnet. 
Minimizing naive proxy objectives for the multi-objective federated optimization doesn't address challenges \textbf{C1, C2}. For instance, a straightforward proxy objective to train the supernet in FL is a linear combination of subnets' losses across all the data partitions. However, minimizing this objective leads to interference: a phenomenon where smaller subnets interfere with larger subnets. The interference results in sub-optimal accuracy of subnets and doesn't solve \textbf{C2}. Moreover, it also leads to slow convergence increasing the training cost (doesn't solve \textbf{C1}).


To efficiently perform multi-objective federated optimization of a large number of subnets, we propose \proposedTraining: an FL training algorithm in \proposedfed that trains supernets with reduced interference for better accuracy (for \textbf{C2}) in a single FL training stage with lower communication/computation costs (for \textbf{C1}). The key idea in \proposedTraining is to optimize a novel objective that explicitly minimizes the loss of worst-performing subnets on each data partition. Optimizing this objective enables \proposedTraining to adapt DNN architectures to different client data distributions. Moreover, improving the worst-performing subnets on each data partition improves performance of best-performing subnets on every data partition due to weight-sharing and reduces interference. To effectively optimize this novel objective, \proposedTraining innovates on subnet sampling and supernet's parameter aggregation.
 In summary, our contributions are as follows: 
\begin{tightitemize}
\item \proposedfed: A one-stage federated NAS method that trains a rich diversity of DNN architectures for efficient on-device inference.
\item \proposedfed produces specialized DNN architectures for $N$ deployment (hardware/latency/MAC) targets 
with O(1) cost. Post the training stage of \proposedfed, the search doesn't require any additional training.
\item \proposedTraining: An FL training algorithm that optimizes a novel objective to train supernets in FL and reduce interference.
\end{tightitemize}

\noindent \proposedfed outperforms existing federated NAS methods across multiple image (CIFAR10/100, CINIC-10) and text datasets (Shakespeare derived from LEAF~\cite{leaf}), degrees of non-iidness, and client participation. It achieves upto $37.7$\% higher accuracy for the same MACs or upto $8.13$x MACs reduction for the same accuracy than existing federated NAS methods with $11$x training cost reduction to satisfy 20 deployment targets. 
\vspace{-2ex}
\section{Related Work}
\vspace{-2ex}
\label{sec:related_work}


Tab.~\ref{tab:rlcomp} compares  \proposedfed with existing FL approaches.


\myparagraph{NAS in FL}  
Existing federated NAS \cite{fednas, fedpnas, yuan2022resource} methods simultaneously search and train DNN architectures, which becomes prohibitively expensive to satisfy multiple inference deployment targets (\figref{fig:overll_dataflow}). \proposedfed decouples the training from search. It only performs training once and enables training-free search to scale to multiple deployment targets. Apart from this key difference, existing federated NAS approaches also differ \wrt DNN architectures.
The architecture search space of FedNAS~\cite{fednas} remains considerably restricted: 
its architecture space does not include DNNs that differ per layer or in depth due to repeated stacking (\secref{sec:intro}). 
FedNASMobile~\cite{yuan2022resource} uses DNN pruning. Its DNN architectures only differ in width and not depth. FedPNas \cite{fedpnas} keeps the base architecture the same among all clients, but supports the DNNs that differ in local layers appended to the base architecture. The common base architecture shared by all DNNs 
restricts architecture diversity.
In contrast, \proposedfed's architecture space allows DNNs to differ at layer granularity by supporting different width per layer and varied number of layers per stage.
The resulting diverse architecture space is essential to produce MACs/latency-efficient DNNs for different hardware (\textbf{C2}). 

\myparagraph{NAS in Centralized Setting}
Recent centralized NAS methods \cite{proxylessnas,mnasnet, fbnet,ofa,compofa} produce DNNs suited for inference. They find the most accurate DNN architectures under latency/FLOPs targets on different hardware. Among these methods, Once-for-all NAS methods like OFA \cite{ofa} and CompOFA \cite{compofa} reduce the training cost to find optimal DNNs by decoupling the training and search stages. In the training stage, OFA jointly optimizes many DNN architectures (subnets) contained inside the supernet with a multi-staged training algorithm.  Once the supernet is trained, OFA performs a search with no additional training to find specialized DNN architectures for target latency/FLOPs. However, all these NAS methods are designed to work in centralized data settings where there are no communication cost restrictions, and techniques like OFA \cite{ofa} can afford to have multi-staged training (doesn't target \textbf{C1}). In contrast, \proposedfed performs NAS in a single-stage FL-training setting and incurs less communication cost.\alexey{weak comparison -- can be made more assertive in the camera ready.}

\myparagraph{System Heterogeneity in FL}  
Many works in FL support system heterogeneity \cite{heterofl,ensemble-distill-fl,yu2022resource, chan2023internal}. These works perform federated optimization under training time constraints like low bandwidth etc. This goal is complementary to \proposedfed's goal that targets satisfying deployment targets at inference (post training). Fundamentally, these works don't perform NAS (as done in \proposedfed): they don't search for optimal DNN architectures based on clients' data distributions and latency/MAC deployment targets at inference. Incorporating training time constraints in \proposedfed is left as future work. 

\begin{table}[tbp]
\centering
\resizebox{\textwidth}{!}{
\begin{tabular}{|l|c|c|c|c|c|c|}
\hline
\textbf{Feature} & \textbf{FedNAS \cite{fednas}} &
\textbf{FedPNAS \cite{fedpnas}} &
\textbf{FedNASMo.~\cite{yuan2022resource}} &  
\textbf{ScaleFL~\cite{yu2022resource}} &  \textbf{InCo~\cite{chan2023internal}} &
\textbf{\proposedfed} \\ \hline
Weight Sharing &  &  &  & &&\usym{1F5F8} \\ \hline
Utilizing NAS & \usym{1F5F8} & \usym{1F5F8} &\usym{1F5F8} &  & & \usym{1F5F8}  \\ \hline
Training Cost for $N$ Deployment & $O(N)$  & $O(N)$ & $O(N)$ & $O(N)$ & $O(N)$ &
$O(1)$ \\ \hline

 Satisfying Diverse Deployment at Inference &  &  &  & &&\usym{1F5F8} \\ \hline
\end{tabular}}
\caption{Comparing existing FL approaches with \proposedfed }
\label{tab:rlcomp}
\end{table}

\vspace{-3ex}
\section{Method}
\label{sec:method}
\vspace{-2ex}
\proposedfed consists of an FL training stage that co-trains many subnets contained inside the supernet cost efficiently (for \textbf{C1}). Once the supernet is globally FL-trained, the clients perform NAS locally to extract optimal subnets (DNN architectures) subject to their diverse deployment targets (for \textbf{C2}) with no additional training. 
Since the local NAS does not require any training and is decoupled, \proposedfed's search is significantly faster than prior federated NAS methods. We begin this section with the description of \proposedfed's training stage including its problem formulation. We explain \proposedTraining: a training technique that co-trains subnets in \proposedfed with reduced interference and compare it with other naive supernet FL-training approaches. Later, we dive deeper into \proposedfed's local NAS stage.

\vspace{-3ex}
\subsection{Problem Formulation}
\label{sec:method:problem_formulation}
\vspace{-2ex}
\myparagraph{\proposedfed's DNN Architecture Space} 
\begin{figure*}[t]
\vspace{-0.7cm}
    \centering
    \begin{subfigure}[b]{0.22\textwidth}
        \includegraphics[height=2cm]{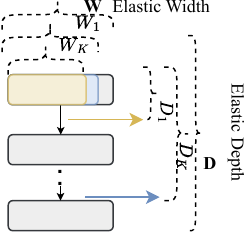} 
        \caption{Supernet}
        \label{fig:WSConcept}
    \end{subfigure}
    \hfill
    \begin{subfigure}[b]{0.37\textwidth}
        \includegraphics[width=\textwidth]{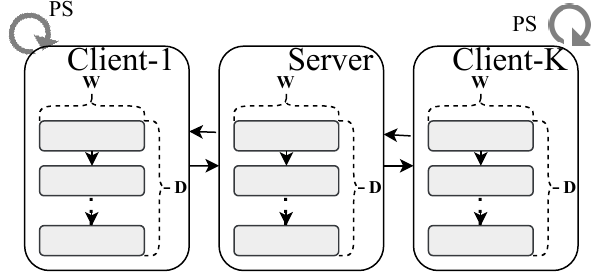}
        \caption{Naive Multi-Stage Supernet FL}
        \label{fig:multi_stage}
    \end{subfigure}
    \hfill
    \begin{subfigure}[b]{0.37\textwidth}
        \includegraphics[width=\textwidth]{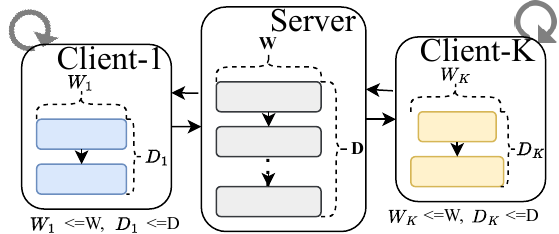}
        \caption{Naive Single-Stage Supernet FL}
        \label{fig:single_stage}
    \end{subfigure}
    \caption{\textbf{Supernet and its Naive FL-Training Methods.} (a) Supernet consists of subnets within it that differ in depth/width. (b) Multi-Stage method sends entire supernet to clients. (c) Single-Stage Supernet FL-Training sends subnets to clients. } 
    \label{fig:overall_dataflow}
     \vspace{-0.7cm}
\end{figure*}
In \proposedfed, DNN architectures differ in depth and width and share their weights as part of a single supernet. These DNN architectures are formally defined by the elastic dimensions of the supernet namely elastic depth and width. Following the common practice of many DNN models \cite{resnet, mobile_bert, mobilenets, densenet}, the supernet consists of multiple stages and each stage consists of multiple blocks. \figref{fig:WSConcept} illustrates a stage in the supernet. Elastic depth decides the number of blocks selected in each stage for a specific DNN architecture. Elastic width decides the width (e.g number of convolution channels) selected in each block. In our experiments, the supernet has four stages, the depth in each stage is chosen from $\{1,2,3\}$, the width expand ratio is chosen from  $\{0.1, 0.14, 0.18, 0.22, 0.25\}$ that roughly equals $(5^1 + 5^2 + 5^3)^4 \approx 5*10^8$ diverse DNN architectures. 
We use $\mathcal{A} = \{\alpha_1, \alpha_2, ..., \alpha_M\}$ to denote the ordered set of $M$ DNN architectures (subnets, $M\approx 5*10^8$ ) \wrt MACs. These subnets share the same weights ($W$) and only require $287.1$ MB for storage. 

\myparagraph{Prior NAS (Supernet Training) Formulation}  With the \proposedfed's architecture space defined, we first elaborate on the prior formulation used in centralized NAS techniques \cite{ofa, compofa} that train supernets in non-federated settings.  
The objective of centralized NAS techniques \cite{ofa} is formalized as follows:
\begin{equation}
\setlength{\belowdisplayskip}{0pt} \setlength{\belowdisplayshortskip}{0pt}
\setlength{\abovedisplayskip}{0pt} \setlength{\abovedisplayshortskip}{0pt}
 \label{eq:objective_ofa}
 \begin{split}
    \min_{W}   \mathop{{}\mathbb{E}}_{\alpha_i \in \mathcal{A}}  \left[ L(\mathcal{G}(W , \alpha_i))\right] \;\; \text{s.t.}\;\; \mathcal{P}(\alpha_i \in \mathcal{A}) = \frac{1}{M}, \; |\mathcal{A}| = M
\end{split}
\end{equation}

\noindent where  $\mathcal{G}(W, \alpha_i)$  denotes a selection (subset) of  $\alpha_i$'s parameters from shared weights $W$. $L(\mathcal{G}(W, \alpha_i))$ denotes the loss of subnet $\alpha_i$ on a central dataset. Centralized NAS techniques minimize the expected loss of all DNN architectures. 

\myparagraph{FL Notation} 
\noindent $K$ clients have their own data partition $P_k$. The size of the partition is denoted by $n_k$ ( where $n=\sum_{k=1}^{K} n_k$ is the total number of data points). $L_k(w) = \sum_{i \in P_k} l_i(w)$  denotes the loss for data points of partition  $P_k$.




\myparagraph{\proposedfed's Training Stage Naive Formulation} \proposedfed's training stage performs the federated optimization of the shared weights ($W$) of the architecture space $\mathcal{A}$. Thus, the objective function of such training is:
\begin{equation}
\setlength{\belowdisplayskip}{0pt} \setlength{\belowdisplayshortskip}{0pt}
\setlength{\abovedisplayskip}{0pt} \setlength{\abovedisplayshortskip}{0pt}
 \label{eq:objective}
 \begin{split}
         \min_{W} \mathop{{}\mathbb{E}}_{\alpha_i \in \mathcal{A}}\left[  \sum_{k=1}^{K} \frac{n_k}{n} * L_k(\mathcal{G}(W , \alpha_i))\right] \;\; \text{s.t.}\;\; \mathcal{P}(\alpha_i \in \mathcal{A}) = \frac{1}{M}
\end{split}
\end{equation}
\objref{eq:objective} finds the weights of the supernet ($W$) that minimize the expected loss of all the DNN architectures in $\mathcal{A}$ on all data partitions $\{ P_1, P_2, ..., P_K \}$. Clearly, \objref{eq:objective} differs from 
\objref{eq:objective_ofa}, and can be viewed as a multi-objective federated optimization with the sub-objective as loss minimization of a subnet on every data partition.  \objref{eq:objective} is different from the objective of personalized FL \cite{pNasFLHyper} that doesn't train DNN architectures globally on all the data partitions.

\vspace{-4ex}
\subsection{Naive Supernet FL-Training Algorithms}
\label{sec:method:naive_supernet_fl}
\vspace{-2ex}
The training algorithms that train the supernet in FL need to optimize \objref{eq:objective}. 
However, optimizing  \objref{eq:objective} is non-trivial due to the expensive expectation over all the DNN architectures' losses on all data partitions. Aggregating gradients of all architectures from all clients is prohibitively expensive and doesn't solve \textbf{C1}, especially for the large architecture space considered in \proposedfed ($\approx 10^{18}$). We first propose two fundamentally distinct naive methods that reformulate \objref{eq:objective} differently to make it tractable:

\myparagraph{Multi-Staged Supernet FL-Training (PS + FL)} 
This method optimizes the following objective:
\begin{equation}
\small
\setlength{\belowdisplayskip}{0.1pt} \setlength{\belowdisplayshortskip}{0.1pt}
\setlength{\abovedisplayskip}{0.1pt} \setlength{\abovedisplayshortskip}{0.1pt}
 \label{eq:objective_ps_fl}
     \min_{W}   \sum_{k=1}^{K} \frac{n_k}{n} * \underbrace{\mathop{{}\mathbb{E}}_{\alpha_i \in \mathcal{A}} \left[ \mathcal{L}_k(\mathcal{G}(W , \alpha_i)\right])}_{\text{Approx. using PS \cite{ofa}}} \;\; \text{s.t.}\;\; \mathcal{P}(\alpha_i \in \mathcal{A}) = \frac{1}{M}
\end{equation}

\noindent \objref{eq:objective_ps_fl} is equivalent to \objref{eq:objective}, however, it is easier to approximate the expectation on a given partition using existing NAS methods. Therefore, this naive method uses OFA's training algorithm Progressive Shrinking (PS) \cite{ofa} and locally runs it in clients to optimize the inner term in \objref{eq:objective_ps_fl} (\figref{fig:multi_stage}). It has been shown in prior works \cite{ofa} that PS effectively approximates the inner term in \objref{eq:objective_ps_fl} by sampling larger subnets initially and gradually sampling smaller subnets in multiple phases.  To adapt PS to the FL setting, we employ multi-stage FL training such that the server provides the entire supernet to each client participating in an FL round along with the PS training phase information (depth, width, etc.). The client trains the supernet locally via PS on its data partition based on the phase provided by the server, updates the supernet's parameters, and sends the locally trained supernet to the server. The server aggregates the supernet parameters from participating clients using the FedAvg \cite{fedavg} algorithm. Overall, we call this method PS+FL as this method adapts  Progressive Shrinking to FL.

\myparagraph{Single-Staged Supernet FL-Training} To achieve supernet's FL-training in a single stage, this method optimizes the following objective:
\begin{equation}
\label{eq:single_stage_fl}
\small
\centering
\begin{split}
\setlength{\belowdisplayskip}{0pt} \setlength{\belowdisplayshortskip}{0pt}
\setlength{\abovedisplayskip}{0pt} \setlength{\abovedisplayshortskip}{0pt}
    \min_{W}  \mathop{{}\mathbb{E}}_{\mathcal{A}_K \subset \mathcal{A}}  \left[ \sum_{\alpha_k \in \mathcal{A}_K}  \frac{n_k}{n} * \mathcal{L}_k(\mathcal{G}(W , \alpha_k))\right] \;\text{s.t.}\; \mathcal{P}(\mathcal{A}_K \subset \mathcal{A}) = \frac{1}{{N\choose k}k!}, \\ 
    \mathcal{A}_K=\{\alpha_i, ..., \alpha_{i+K}\}, \; \forall  \alpha_i,\alpha_j \in \mathcal{A}_K\; \alpha_i \neq \alpha_j, \;    |\mathcal{A}_K| = K
\end{split}
\end{equation}
\objref{eq:single_stage_fl} minimizes the expected loss of any K ordered DNN architectures selected from $\mathcal{A}$ (M DNN architectures) mapped to their specific data partition ($a_k$ is uniquely mapped to $L_k$). Note that the probability of selecting K-ordered DNN architectures is uniform \wrt all permutations ($\mathcal{P}(\mathcal{A}_K \subset \mathcal{A}) = \frac{1}{{N\choose k}k!}$).
Therefore, in this method, the server uniformly samples $K$\footnote{the method can sample DNN architectures less than $K$, say $C*K$ if the client participation ratio is $C < 1$.} DNN architectures from the architecture space $\mathcal{A}$. Then, it randomly assigns the sampled architectures (subnets) to each participating client and sends subnets' partial parameters ($\mathcal{G}(W, \alpha_k)$), illustrated in \figref{fig:single_stage}. On receiving a specific subnet from the server, the client trains the subnet locally on its data partition and sends it back to the server. The server receives different subnets that vary in shape and size from different clients. The subnets' parameters partially overlap with each other due to weight-sharing. Therefore, the server performs cardinal averaging: for each supernet parameter, only the clients' subnets that share that parameter are averaged.



\begin{table}[tb]
\centering
\resizebox{0.9\columnwidth}{!}{
\small
    \begin{tabular}{|c|c|c|c|c|}
        \hline
        \multirow{2}{*}{\small{DNN Arch.$\in \mathcal{A}$}}&\multirow{2}{*}{Method}&\multicolumn{3}{c}{Test Accuracy (\%)}\vline\\ \cline{3-5}
        &&non-iid=100 & non-iid=1 & non-iid=0.1 \\ \hline
        \multirow{4}{*}{Smallest}
            &FedAvg&$85.25\pm0.46$&$83.42\pm 0.19$&$77.15\pm 2.5$\\
            &Single-Staged Supernet FL&$84.6\pm 0.19$&$83.17\pm 0.12$&$76.28\pm 1.31$\\
            &Multi-Staged Supernet FL&$84.53\pm 0.58$&$82.82\pm 0.34$&$76.26\pm 2.35$\\
            &\proposedTraining&$\pmb{89.42\pm 0.11}$&$\pmb{88.69\pm 0.2}$&$\pmb{81.81\pm 1.59}$\\ \hline
        \multirow{4}{*}{Largest}
            &FedAvg&$89.44\pm0.67$&$87.88\pm 0.7$&$81.24\pm 1.99$\\
            &Single-Staged Supernet FL&$87.14\pm 0.2$&$86.03\pm 0.26$&$80.02\pm 2.07$\\
            &Multi-Staged Supernet FL&$86.45\pm 0.53$&$85.02\pm 0.32$&$78.57\pm 2.48$\\
            &\proposedTraining&$\pmb{91.34\pm 0.3}$&$\pmb{90.91\pm 0.15}$&$\pmb{84.72\pm 1.78}$\\ \hline
    \end{tabular}
    }
    \caption{ \small \textbf{Naive Supernet FL-Training Accuracy Comparison.} Test Accuracy compared on CIFAR10 dataset paritioned with different levels of non-iidness among 20 clients, 40\% client-participation. \proposedTraining outperforms the naive supernet FL-Training methods (single/multi). The naive methods are inferior to FedAvg.}
    \label{tab:naive_comparison}
    \vspace{-1cm}
\end{table}
\begin{figure}[tb]
    \vspace{-0.8cm}
    \includegraphics[width=\columnwidth]{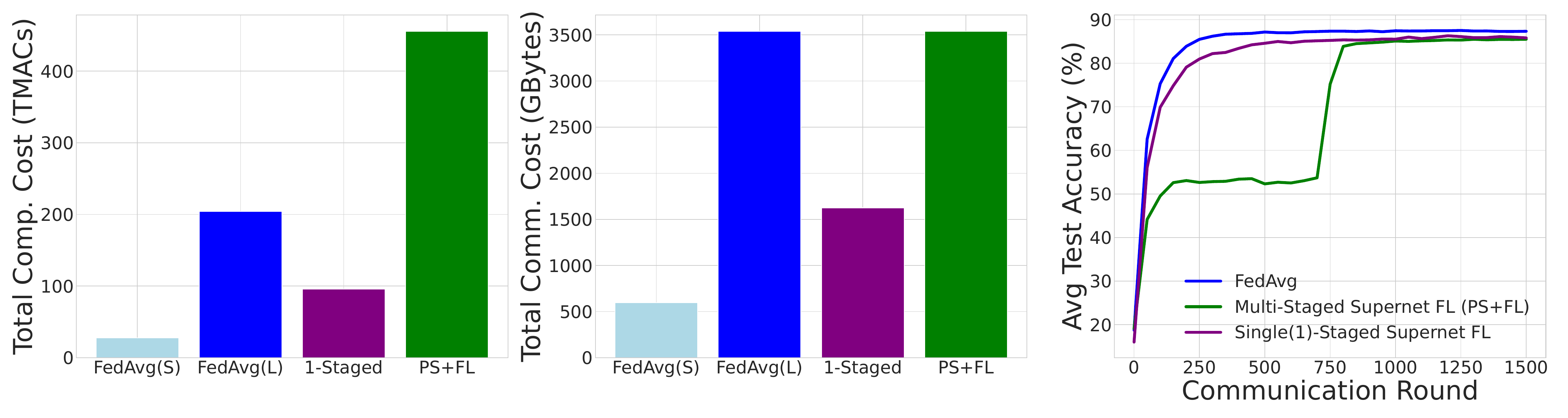}
    \vspace{-3ex}
    \caption{\textbf{Naive Supernet FL-Training Cost/Convergence Comparison.} Communication/Computational cost (left,middle) are compared for naive supernet FL approaches with FedAvg training of largest/smallest subnet. The right plot compares convergence over 1500 rounds. Naive-methods have slow convergence and high training cost.}
    \label{fig:naive_comparison:cost_convergence}
    \vspace{-0.8cm}
\end{figure}

\noindent \myparagraph{Comparing Naive Supernet FL-Training Methods} We compare the two naive federated supernet training methods. To meaningfully understand the accuracy gaps, the naive methods are also compared with FedAvg \cite{fedavg} which trains the smallest and largest subnets (no weight-sharing). The smallest/largest subnets in $\mathcal{A}$ represent the lower and upper bound of accuracy reached by all DNNs in  $\mathcal{A}$. \tabref{tab:naive_comparison} describes the FL setting.

\myparagraph{Training Costs Comparison} The takeaways \wrt training cost are as follows:
\begin{tightitemize}

    \item The multi-stage naive method incurs more communication/computation cost than the single-stage naive method (left, middle plot in \figref{fig:naive_comparison:cost_convergence}). 
    This is because the multi-stage method sends the entire supernet ($|W|$) back and forth to each participating client in each round, whereas, the single-stage method only sends partial weights of the supernet (in the form of subnets) to each client that leads to communication cost savings ($|\mathcal{G}(W , \alpha_{i})| \leq |W|$). The multi-stage method has more computational cost as PS that runs locally samples more than one architecture in each iteration 
    (PS \cite{ofa} samples 4 subnets in each minibatch), while, single-stage method trains only one architecture locally (\figref{fig:overall_dataflow}). 
    \item Multi-stage method's communication cost is the same as the cost of training the largest network using FedAvg. This is because the largest subnet subsumes all parameters of the supernet ($ \max_{\alpha_i} |\mathcal{G}(W, \alpha_{i})| = W$) and clients receive all parameters of the supernet in both the methods. Single-stage naive method's communication cost lies between the cost of FedAvg training of the smallest/largest subnet as partial parameters are sent to clients in each round ($ \min_{\alpha_i} |\mathcal{G}(W, \alpha_{i})| < 1/M * \sum_{\alpha_i}|\mathcal{G}(W, \alpha_{i})| < \max_{\alpha_i} |\mathcal{G}(W, \alpha_{i})|$).    
\end{tightitemize}
\myparagraph{Accuracy Comparison} As seen in \tabref{tab:naive_comparison}, both the naive supernet FL-training methods fail to match the test accuracy of the smallest or largest subnets trained without weight sharing using FedAvg. Hence, the naive methods don't provide optimal DNN architectures (don't solve \textbf{C2}). We attribute the following reasons to the sub-optimal accuracy of naive supernet FL-training methods:
\begin{tightitemize}
\item \textit{Interference.} The accuracy of the largest subnet in naive supernet FL-training methods is sub-par accuracy compared to FedAvg. 
We argue this is because of interference: phenomena also observed in centralized supernet training methods \cite{ofa,bignas} where smaller subnets affect the accuracy of larger subnets. 

\item \textit{Slow Convergence.} Both the naive methods converge slowly compared to FedAvg as seen in \figref{fig:naive_comparison:cost_convergence} (right plot). The multi-stage method suffers from slow convergence due to the multi-phase approach. It introduces smaller subnets progressively which makes the convergence of smaller subnets slow. The naive single-stage method suffers from slow convergence as all the supernet's parameters ($W$) may not get updated in every FL round, leading to staleness. This is because the method that optimizes \objref{eq:single_stage_fl} randomly samples $K$ architectures in each round, which may not span the entire supernet parameters \ie $\exists \mathcal{A}_K \;s.t.\; \bigcup_{\alpha_k \in \mathcal{A}_K}\mathcal{G}(W, \alpha_{k}) \neq W$. 

\end{tightitemize}

\vspace{-2ex}
\subsection{\proposedTraining: Proposed Supernet FL-Training Algorithm}
\label{sec:method:proposed}
\vspace{-1ex}
\proposedTraining is a supernet FL training algorithm that produces optimal DNN architectures (for \textbf{C2}) and trains the supernet in a single stage. The single-stage training offers lower training cost (for \textbf{C1}) compared to multi-stage supernet FL training methods (\secref{sec:method:naive_supernet_fl}). \proposedTraining produces optimal DNN architectures as it optimizes a fundamentally different objective to reduce interference:


{
\footnotesize
\begin{equation}
\small
\label{eq:proposed_stage_fl}
\setlength{\belowdisplayskip}{0pt} \setlength{\belowdisplayshortskip}{0pt}
\setlength{\abovedisplayskip}{0pt} \setlength{\abovedisplayshortskip}{0pt}
\centering
\begin{split}
    \min_{W}  \max_{\overrightarrow{\gamma}}\mathop{{}\mathbb{E}}_{\alpha_i \in \mathcal{A}}  \left[ \sum_{k=1}^{K} \gamma_{ik} * \frac{n_k}{n} * \mathcal{L}_k(\mathcal{G}(W , \alpha_i))\right] \text{ s.t. } \mathcal{P}(\alpha_i \in \mathcal{A}) = \frac{|\mathcal{G}(W , \alpha_i)|}{\sum_{i=1}^M|\mathcal{G}(W , \alpha_i)|}, \\ 
    \forall i \in [1,M],k \in [1,K]: \gamma_{ik} \in \{0,1\} , \sum_{i=1}^M \gamma_{ik} = 1, \sum_{i=1}^M\sum_{k=1}^K \gamma_{ik} = K
\end{split}
\end{equation}
}


\noindent \objref{eq:proposed_stage_fl} minimizes the expected loss of worst-performing DNN architectures on each data partition. $\gamma_{ik}$ is an indicator that represents the DNN architecture $\alpha_i$ evaluated on data partition $P_k$. The maximum over $\overrightarrow{\gamma}$ picks worst-performing DNN architectures (maximum loss) on each data partition. The constraint $\sum_{i=1}^M \gamma_{ik} = 1$ ensures that only one architecture gets assigned to $k^{th}$ data partition. Therefore, in \proposedTraining, each client (data partition) receives one DNN architecture (in expectation) similar to single-staged method in \secref{sec:method:naive_supernet_fl}. Overall, \objref{eq:proposed_stage_fl} has two key features:
\begin{tightitemize}
      \item \textit{Improving Worst-Performing DNN Architectures on each Data Partition.} 
      The key insight of \proposedTraining is that improving accuracy (by minimizing loss) of worst-performing DNN architectures on each data partition has a potential to improve the accuracy of best performing DNN architectures across all the data partitions as the weights are shared. This optimization is particularly helpful in non-iid settings where each data partitions differs significantly. Due to the data heterogeneity, arbitrarily different DNN architectures may perform worse on different partition. Under such settings, \objref{eq:proposed_stage_fl} enables DNN architectures to adapt to different data distributions by explicitly minimizing the loss of worst performing DNN architectures on each data partition.
      
      
    \item \textit{Weight-Shared Based Sampling Probability.} Instead of uniform sampling, \objref{eq:proposed_stage_fl} prioritizes sampling the DNN architectures that share their weights most with the supernet ($\mathcal{P}(\alpha_i \in \mathcal{A}) = \frac{|\mathcal{G}(W , \alpha_i)|}{\sum_{i=1}^M|\mathcal{G}(W , \alpha_i)|}$). This ensures that most of the supernet's parameters get updated as part of the optimization, mitigating the staleness observed in naive single-staged supernet FL-training method (\secref{sec:method:naive_supernet_fl}).
  
\end{tightitemize}
However, optimizing \objref{eq:proposed_stage_fl} remains prohibitively expensive due to \proposedfed's large architecture space (\secref{sec:method:problem_formulation}). To make the optimization tractable, \proposedTraining innovates on DNN architecture sampling and parameter aggregation:

\myparagraph{\proposedTraining's Subnet Sampling} \proposedTraining approximates subnet selection ($\alpha_i$) in \objref{eq:proposed_stage_fl}. While, $\max_{\vec{\gamma}}$ enables selection of smaller subnets as smaller subnets typically have high loss, the weight-shared based probability  prioritizes the sampling of larger subnets. To approximate this, \proposedTraining samples the largest, smallest and random subnets (uniformly) in $\mathcal{A}$ in each FL round. The largest and smallest subnets explicitly sampled in \proposedTraining act as lower/upper bounds of accuracy.  Optimizing both the bounds improves performance of other subnets due to weight-sharing. To improve worst-performing DNN architectures on different data partitions, \proposedTraining gives the smallest/largest subnets to the clients that have received these subnets the least in each FL-round. \proposedTraining approximates $\max_{\gamma}$ in \objref{eq:proposed_stage_fl} as the clients that receive largest/smallest subnets the least have high loss \wrt these subnets on their data partitions.

\myparagraph{\proposedTraining's Parameter Aggregation} The parameter aggregation in \proposedTraining is based on the optimization dynamics of \objref{eq:proposed_stage_fl}. Initially, when the loss of all the subnets in $\mathcal{A}$ is roughly equal, larger subnets get optimized due to weight-shared based probability. As the optimization progresses, the smaller subnets have higher loss and get optimized due to $\max_{\vec{\gamma}}$. To emulate this optimization dynamics, \proposedTraining performs weighted parameter aggregation. It assigns a weight of $\beta \in (0,1)$ to largest subnet's parameters and $(1-\beta)$ to the parameters of rest of the subnets. Initially, $\beta$ is assigned a high value ($\beta_0 = 0.9$) to prioritize updates from the largest subnet. As FL-rounds progress, $\beta$ is decayed to make smaller subnets contribute more in the supernet FL training. We provide an extensive evaluation on both the initial $\beta$-value and its decay function in \secref{sec:experiments}. 

\noindent \proposedTraining's performance is shown in \tabref{tab:naive_comparison}. \proposedTraining achieves superior performance compared to the naive supernet FL training methods even under extreme non-iidness (0.1). \proposedTraining benefits from optimizing \objref{eq:proposed_stage_fl} and improving worst-performing subnets on each data partition (helpful in non-iid settings).
\vspace{-4ex}
\subsection{\proposedfed's Search Stage}
\label{sec:method:search}
\vspace{-2ex}
Once the supernet is trained in FL using \proposedTraining, \proposedfed's search stage finds the optimal DNN architectures subject to client's deployment targets (\textbf{C2}). \proposedfed's search stage is formulated as a constraint optimization problem:

\begin{equation}
\label{eq:proposed_search}
\small
\setlength{\belowdisplayskip}{0pt} \setlength{\belowdisplayshortskip}{0pt}
\setlength{\abovedisplayskip}{0pt} \setlength{\abovedisplayshortskip}{0pt}
\centering
\begin{split}
    \min_{\alpha \in \mathcal{A}} L_{val}(\mathcal{G}(W^{*} , \alpha_i)) \;\;s.t. \;\;\text{MACs}(\alpha) = \theta
\end{split}
\end{equation}
\noindent $W^{*}$ is supernet's parameters trained using \proposedTraining and $\theta$ is the MAC (Multiply-and-Add) constraint (or any other depl. target). $L_{val}$ represents loss on either a local or global (if available) validation dataset. Finding optimal DNN architecture ($\alpha_*$) doesn't require any re-training. Optimizing \objref{eq:proposed_search} is fast as it simply involves evaluating subnets for accuracy/FLOPs. 
Applying \objref{eq:proposed_search} to multiple deployment targets ($\theta_i$'s) doesn't add any training cost. This makes \proposedTraining O(1) \wrt deployment targets. To speed-up the search, \proposedfed uses predictors (three layer MLPs) for estimating accuracy and latency of subnets, similar to OFA\cite{ofa}. \proposedfed uses predictor-guided evolutionary search \cite{predictor_guided_search} to optimize \objref{eq:proposed_search}.  
 


\vspace{-3ex}
\section{Experiments}
\label{sec:experiments}
\vspace{-2ex}
We evaluate whether \proposedfed produces efficient DNN architectures (solves \textbf{C2}, \secref{sec:intro}) and generalizes on various FL scenarios: (1) real-world text/image datasets (2) non-iidness (3) client participation ratio ($C$) under multiple MAC targets. We also analyze \proposedfed's scalability to satisfy multiple deployment scenarios by reporting its training cost (\textbf{C1}, \secref{sec:intro}). We perform a detailed ablation study on \proposedTraining's hyper-parameters and its ability to specialize DNN architectures on different hardware/latency targets. 

\vspace{-2ex}
\subsection{Setup}
\label{sec:experiment:setup}
\vspace{-2ex}

\noindent \textbf{Baselines.} We compare \proposedfed with FedAvg \cite{fedavg} and existing federated NAS methods: FedNAS \cite{fednas}, FedPNAS \cite{fedpnas}. The comparisons are done over four deployment targets\footnote{FedPNAS could only satisfy two deployment targets at lower MACs} (Multiply-and-Additions (MACs) constraints). The baselines are run repeatedly for each deployment target, \proposedfed is run only once.

\noindent \textbf{Dataset and Models.} Experiments are done on three images and one text datasets: CIFAR10/100 \citep{cifar}, CINIC-10 \citep{cinic10}, and Shakespeare derived from LEAF benchmark\footnote{The data is partitioned based on each role in a play in non-IID setting} \cite{leaf}. For image datasets, \proposedfed's supernet is a ResNet \citep{resnet} based architecture, containing ResNet-10/26 as the smallest/largest subnets respectively. For the text dataset, the base supernetwork is a TCN \citep{tcn} based architecture. For FedAvg, the DNNs are manually chosen following the scaling rule in \cite{efficient_net}. The DNN-archs of FedNAS, FedPNAS are kept the same as described in their method. The validation dataset is used for the search phase in all federated NAS methods including \proposedfed, the setting is the same as FedNAS \cite{fednas}.

\noindent \textbf{Non-iid Setting.} Similar to \citep{fed_distill, ensemble-distill-fl}, we use  the Dirichlet distribution for non-iid training data. The non-iid degree=100 is close to the uniform distribution of classes.  Lower non-iid degree (like 0.1) increases per-class differences in data.

\noindent \textbf{Training Hyper-params.} For image datasets, \proposedfed's setting is similar to \citep{ensemble-distill-fl} using local SGD \citep{local-sgd} with no weight decay and a constant learning rate.

\vspace{-3ex}
\subsection{Comparison with Baselines}
\label{sec:experiment:eval}
\vspace{-1ex}

\myparagraph{Comparison on Image Datasets}  \tabref{tab:NAS_comparison_image_dataset} compares \proposedfed with the baselines on three image datasets. For this experiment, we keep the client participation as $C=0.4$ and non-iid degree=100. CINIC10/CIFAR10/100 are run for $R=1000/1500/2000$ communication rounds. CINIC10 is divided into K=100 partitions (\#. clients) and CIFAR10/100 are divided into K=20 partitions.

\noindent \textit{Takeaway}. \proposedfed achieves \textbf{upto 13.1\% more accuracy} for target MACs. It finds better DNN architectures across multiple MAC targets compared to the baselines on different datasets.  Specifically, \proposedfed outperforms the baselines at lower MACs and tougher dataset (CIFAR100). \proposedfed's superior performance comes from co-training a large number of diverse DNN architectures. This enables the search for DNN architectures with arbitrary width and depth to satisfy MAC targets. This DNN architecture diversity remains restrictive in existing federated NAS methods and leads to sub-optimal performance. 

\begin{table}[tb]
\centering
\begin{minipage}[t]{0.54\textwidth}
    \centering
    \captionsetup{width=0.9\textwidth}
    \resizebox{\textwidth}{!}{
    \footnotesize
    \begin{tabular}{|c|c|c|c|c|}
        \hline
        \multirow{2}{*}{Billion MACs}&\multirow{2}{*}{Method}&\multicolumn{3}{c}{Test Accuracy (\%)}\vline\\ \cline{3-5}
        &&CIFAR10 & CIFAR100 & CINIC10 \\ \hline
        \multirow{4}{*}{0.45-0.95}
            &FedAvg&$85.25\pm0.46$&$43.19\pm 0.54$&$61.76\pm 0.78$\\
            &FedNAS&$77.33\pm 0.31$&$40.92\pm 2.21$&$58.15\pm 0.18$\\
            &FedPNAS&$88.83\pm 0.5$&$45.77\pm 0.68$&$64.3\pm 0.98$\\
            &\proposedfed&$\pmb{89.42\pm 0.11}$&$\pmb{56.35\pm 0.3}$&$\pmb{73.12\pm 0.77}$\\ \hline
        \multirow{3}{*}{0.95-1.45}
            &FedAvg&$86.36\pm 0.22$&$43.92\pm 0.57$&$63\pm 0.17$\\
            &FedPNAS&$89.27\pm 0.51$&$47.8\pm 26$&$66.74\pm 0.32$\\
            &\proposedfed&$\pmb{90.22\pm 0.31}$&$\pmb{57.16\pm 0.23}$& $\pmb{74.5\pm 0.74}$\\ \hline
        \multirow{3}{*}{1.45-2.45}
            &FedAvg&$87.59\pm 0.27$&$44.4\pm 0.56$&$64\pm 0.07$\\
            &FedNAS&$86.41\pm 0.1$&$55.82\pm 0.29$&$69.97\pm 0.27$\\
            &\proposedfed&$\pmb{90.93\pm 0.23}$&$\pmb{57.85\pm 0.31}$& $\pmb{75.08\pm 0.7}$\\ \hline
        \multirow{3}{*}{2.45-3.75}
            &FedAvg&$89.44\pm0.67$&$45\pm 0.27$&$66.02\pm 0.13$\\
            &FedNAS&$89.43\pm 0.36$&$58.39\pm 0.23$&$71.93\pm 0.13$\\
            &\proposedfed&$\pmb{91.34\pm 0.3}$&$\pmb{58.25\pm 0.39}$&$\pmb{75.38\pm 0.73}$\\ \hline
    \end{tabular}   
    }
    \caption{\scriptsize {\textbf{Image Datasets Comparison.} \proposedfed compared with FedAvg, FL-NAS methods on image datasets for different MAC targets. \proposedfed consistently outperforms the baselines.}}
    \label{tab:NAS_comparison_image_dataset}
\end{minipage}%
\begin{minipage}[t]{0.46\textwidth}
    \centering
    \captionsetup{width=0.9\textwidth}
    \resizebox{\textwidth}{!}{
    \footnotesize
    \begin{tabular}{|c|c|c|}
        \hline
        Million MACs&Method& Test Accuracy (\%)\\ \hline
        \multirow{2}{*}{0-0.5}&FedAvg&$48.52\pm 0.11$\\
            &\proposedfed&$48.22\pm0.27$\\ \hline
        \multirow{2}{*}{0.5-1}&FedAvg&$49.17\pm 0.02$\\
            &\proposedfed&$\pmb{49.81\pm0.16}$\\ \hline
        \multirow{2}{*}{1-1.5}&FedAvg&$51.94\pm 0.03$\\
            &\proposedfed&$\pmb{53.26\pm0.06}$\\ \hline
        \multirow{2}{*}{1.5-2.75}&FedAvg&$53.48\pm 0.09$\\
            &\proposedfed&$\pmb{54.59\pm0.15}$\\ \hline
        \multirow{2}{*}{2.75-4.0}&FedAvg&$53.62\pm 0.1$\\
            &\proposedfed&$\pmb{54.61\pm0.13}$\\ \hline
    \end{tabular}
    }
    \caption{\scriptsize{\textbf{Text Dataset Comparison.} Comparison on tough FL setting: Shakespeare dataset \cite{leaf}, C=4\%, non-iidness, K=660 partitions. \proposedfed finds efficient DNN archs under tough FL setting.}}
    \label{tab:Shakespeare_dataset}
\end{minipage}
\vspace{-0.8cm}
\end{table}


\begin{table}[tb]
\centering
\begin{minipage}[t]{0.55\textwidth}
    \centering
    \captionsetup{width=0.9\textwidth}
    \resizebox{\textwidth}{!}{
    \begin{tabular}{|c|c|c|c|c|}
        \hline
        \multirow{2}{*}{Billion MACs}&\multirow{2}{*}{Method}&\multicolumn{3}{c}{Test Accuracy (\%)}\vline\\ \cline{3-5}
        &&non-iid=100 & non-iid=1 & non-iid=0.1 \\ \hline
        \multirow{4}{*}{0.45-0.95}
            &FedAvg&$85.25\pm0.46$&$83.42\pm 0.19$&$77.15\pm 2.5$\\
            &FedNAS&$77.33\pm 0.31$&$71.38\pm 0.37$&$51.57\pm 3.32$\\
            &FedPNAS&$88.83\pm 0.5$&$85.7\pm 0.4$&$78.73\pm 0.45$\\
            &\proposedfed &$\pmb{89.42\pm 0.11}$&$\pmb{88.69\pm 0.2}$&$\pmb{81.81\pm 1.59}$\\ \hline
        \multirow{3}{*}{0.95-1.45}
            &FedAvg&$86.36\pm 0.22$&$84.65\pm 0.11$&$77.99\pm 1.6$\\
            &FedPNAS&$89.27\pm 0.51$&$87.53\pm 0.32$&$81.13\pm 0.4$\\
            &\proposedfed&$\pmb{90.22\pm 0.31}$&$\pmb{89.3\pm 0.35}$& $\pmb{83.27\pm 1.28}$\\ \hline
        \multirow{3}{*}{1.45-2.45}
            &FedAvg&$87.59\pm 0.27$&$86.14\pm 0.23$&$79.93\pm 1.34$\\
            &FedNAS&$86.41\pm 0.1$&$82.13\pm 0.65$&$65.03\pm 2.57$\\
            &\proposedfed&$\pmb{90.93\pm 0.23}$&$\pmb{90.36\pm 0.21}$&$\pmb{84.1\pm 1.71}$\\ \hline
        \multirow{3}{*}{2.45-3.75}
            &FedAvg&$89.44\pm0.67$&$87.88\pm 0.7$&$81.24\pm 1.99$\\
            &FedNAS&$89.43\pm 0.36$&$85.85\pm 0.35$&$68.13\pm 5.04$\\
            &\proposedfed&$\pmb{91.34\pm 0.3}$&$\pmb{90.91\pm 0.15}$&$\pmb{84.72\pm 1.78}$\\ \hline
    \end{tabular}
    }
    \caption{\scriptsize \textbf{Non-iidness Comparison.} Comparison across varying non-iidness on CIFAR10. \proposedfed outperforms baselines as it adapts DNN archs to non-iidness due to optimization of \objref{eq:proposed_stage_fl}.}
    \label{tab:NAS_comparison_alphas}
\end{minipage}%
\begin{minipage}[t]{0.45\textwidth}
    \centering
    \captionsetup{width=0.9\textwidth}
    \resizebox{\textwidth}{!}{
    \begin{tabular}{|c|c|c|c|}
        \hline
        \multirow{2}{*}{Billion MACs}&\multirow{2}{*}{Method}&\multicolumn{2}{c}{Test Accuracy (\%)}\vline\\ \cline{3-4}
        &&C=0.2 & C=0.4 \\ \hline
        \multirow{4}{*}{0.45-0.95}
            &FedAvg&$85.59\pm 0.59$&$85.25\pm0.46$\\
            &FedNAS&$76.23\pm 0.5$&$77.33\pm 0.31$\\
            &FedPNAS&$86.63\pm 0.51$&$88.83\pm 0.5$\\
            &\proposedfed&$\pmb{89.58\pm 0.5}$&$\pmb{89.42\pm 0.11}$\\ \hline
        \multirow{3}{*}{0.95-1.45}
            &FedAvg&$87.01\pm 0.24$&$86.36\pm 0.22$\\
            &FedPNAS&$87.83\pm 0.21$&$89.27\pm 0.51$\\
            &\proposedfed&$\pmb{89.95\pm 0.57}$&$\pmb{90.22\pm 0.31}$\\ \hline
        \multirow{3}{*}{1.45-2.45}
            &FedAvg&$88.04\pm 0.31$&$87.59\pm 0.27$\\
            &FedNAS&$84.65\pm 0.14$&$86.41\pm 0.1$\\
            &\proposedfed&$\pmb{90.7\pm 0.48}$&$\pmb{90.93\pm 0.23}$\\ \hline
        \multirow{3}{*}{2.45-3.75}
            &FedAvg&$89.96\pm 0.65$&$89.44\pm0.67$\\
            &FedNAS&$88\pm 0.38$&$89.43\pm 0.36$\\
            &\proposedfed&$\pmb{91.16\pm 0.45}$&$\pmb{91.34\pm 0.3}$\\ \hline
    \end{tabular}
    }
    \caption{\scriptsize \textbf{Client Participation.} Comparison \wrt C=0.2,0.4 on CIFAR10 dataset. \proposedfed outperforms baselines. \proposedTraining's subnet sampling effectively optimizes \objref{eq:proposed_stage_fl} at low C.} 
    \label{tab:NAS_comparison_client_participation}
\end{minipage}
\vspace{-0.8cm}
\end{table}

\myparagraph{Comparison on Text Dataset}  We evaluate \proposedfed on a tough FL setting: text dataset, non-iidness, large number of clients (data partitions) K=660, and 4\% client participation. \tabref{tab:Shakespeare_dataset} compares \proposedfed with FedAvg on Shakespeare dataset derived from LEAF \cite{leaf}\footnote{FedNAS, FedPNAS only release their DNN architectures for image datasets.}. 

\noindent \textit{Takeaway.} \proposedfed outperforms FedAvg on shakespeare dataset, is upto \textbf{1.29\%} more accurate. Even under tough FL settings, \proposedfed benefits from automating the design and training of DNN architectures (\secref{sec:intro}). 

\myparagraph{Comparison on Non-iidness} We evaluate if \proposedfed adapts to different data distributions in FL. \tabref{tab:NAS_comparison_alphas} compares \proposedfed with the baselines on different degrees of non-iidness (0.1, 1, 100). We use CIFAR10 dataset for this experiment divided into K=20 partitions with 40\% client participation. 

\noindent \textit{Takeaway.} \proposedfed outperforms the baselines on varied degrees of non-iidness at multiple MAC targets. Particularly, the superiority of \proposedfed \wrt accuracy is more at extreme non-iidness (0.1). This is because  \proposedfed benefits from adapting its DNN architectures to different data distributions by optimizing \objref{eq:proposed_stage_fl} and explicitly improving the worst-performing subnets (\secref{sec:method:proposed}). 

\myparagraph{Effect of Client Participation} We evaluate the efficacy of \proposedfed under different client participation in FL (C=0.2,0.4) on CIFAR10 divided into K=20 partitions. 
We intend to establish whether \proposedTraining can approximate \objref{eq:proposed_stage_fl} under low client participation. \tabref{tab:NAS_comparison_client_participation} compares \proposedfed with the baselines.

\noindent \textit{Takeaway.} \proposedfed achieves better accuracy than the baselines for different MAC targets even under low client participation (20\%). This is because \proposedTraining's subnet sampling effectively approximates  \objref{eq:proposed_stage_fl} under different client participation.
It sends the smallest/largest subnets to the participating clients that have received these subnets the least (\secref{sec:method:proposed}).

\begin{wrapfigure}{r}{0.4\columnwidth}
\vspace{-0.9cm}
 \centering
   \includegraphics[width=0.4\columnwidth]{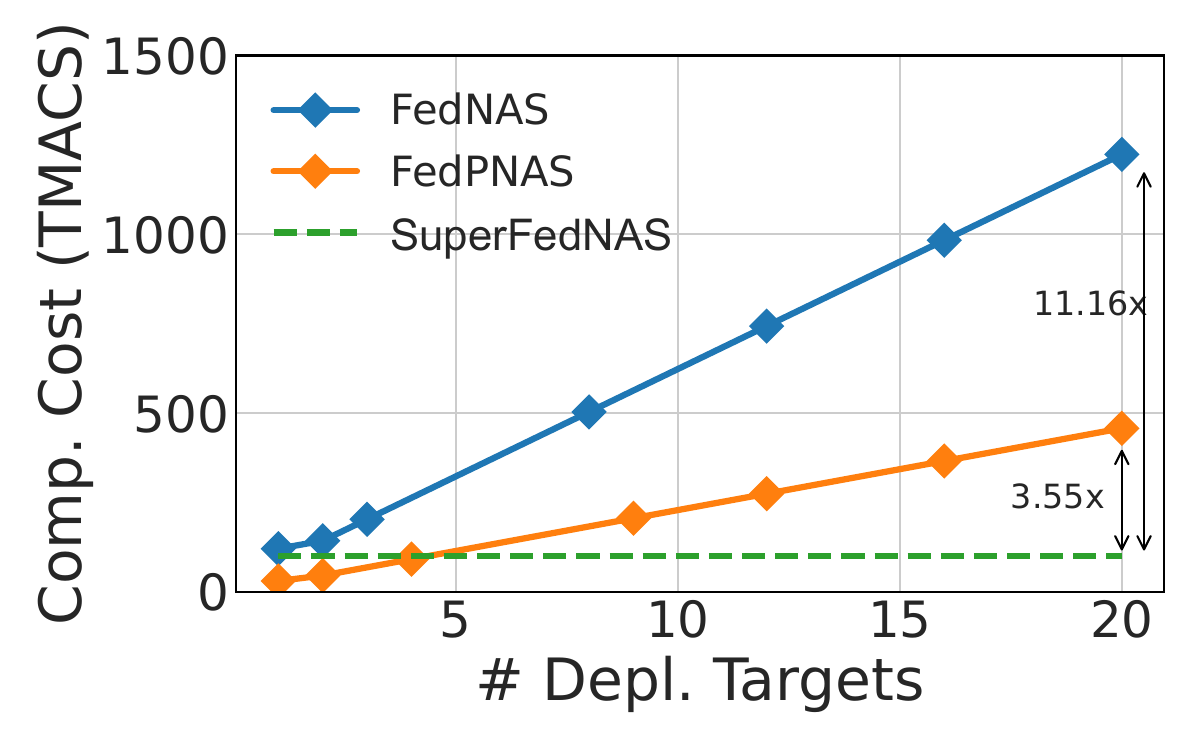}
   \caption{\textbf{Training Cost Comparison.}}
   \label{fig:exp:cost_comparison}
   \vspace{-0.8cm}
\end{wrapfigure}

\myparagraph{Training Cost Comparison} We assess whether \proposedfed is scalable to multiple deployment (solves \textbf{C1}, \secref{sec:intro}).  We report the \proposedfed's training cost to satisfy multiple deployment targets.  \figref{fig:exp:cost_comparison} compares \proposedfed's training cost with  baselines  \wrt  \#. computations required to satisfy 20 depl. targets.  
\noindent \textit{Takeaway.}  \proposedfed's training cost remains O(1) \wrt number of deployment targets and is upto \textbf{6x less} than the baselines. This is because \proposedfed decouples training from the search in federated NAS, enables search without additional training (\secref{sec:method:search}). In contrast, both FedNAS and FedPNAS are run repeatedly to satisfy multiple deployment targets as they train and search simultaneously. This leads to a linear increase in training cost with deployment targets for the baselines.

\vspace{-2ex}
\begin{figure*}[t]
    \centering
    \begin{subfigure}[b]{0.35\textwidth}
        \includegraphics[width=\textwidth]{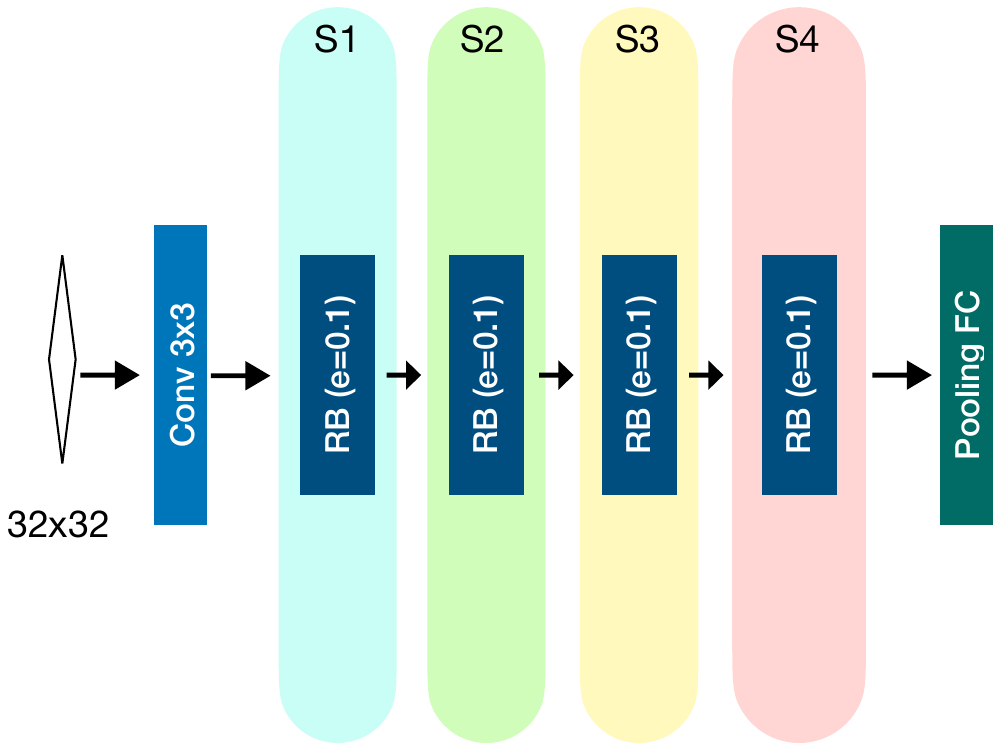} 
        \caption{112 ms lat. target on AMD CPU}
        \label{fig:nas_vis:cpu}
    \end{subfigure}
    \hfill
    \begin{subfigure}[b]{0.55\textwidth}
        \includegraphics[width=\textwidth]{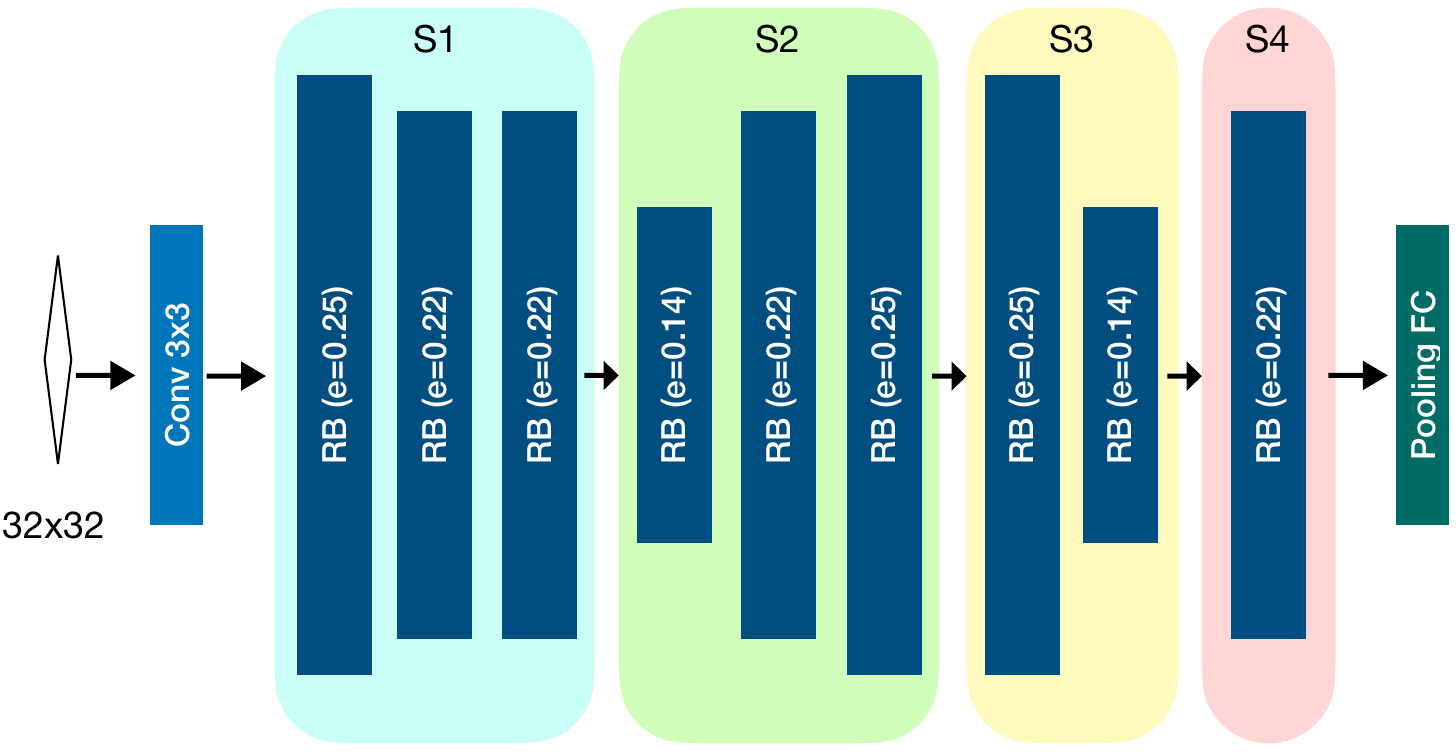}
        \caption{18.2 ms lat. target on Nvidia RTX 2080 Ti}
        \label{fig:gpu}
    \end{subfigure}
    \vspace{-2ex}
    \caption{\small \textbf{\proposedfed's DNN Arch. Specialization.}  Specialized DNNs found by \proposedfed's search stage on different hardware/latency targets. \proposedfed finds a more accurate DNN for RTX 2080Ti GPU (91.56\%) compared to AMD CPU (85.25\%). \proposedfed's specialized DNNs are: shallow/thin for AMD CPU, wide/deep  for GPU.}  
    \label{fig:nas_viz}
     \vspace{-3ex}
\end{figure*}

\subsection{Ablation Study}
\label{sec:experiment:ablation}
\myparagraph{Specialized DNNs for Target Inference Deployments}
We evaluate \proposedfed's ability to specialize DNNs for diverse inference deployment targets. We use the supernet trained on CIFAR10 dataset divided into $k=20$ partitions (an experiment in  \tabref{tab:NAS_comparison_image_dataset}). We use \proposedfed's search stage (\secref{sec:method:search}) to find specialized subnets on two different hardware/latency targets: 18.2 ms on RTX 2080Ti GPU and 112 ms on an AMD CPU. As the search doesn't require re-training, we create a dataset by sampling different subnets and getting their accuracy and latency. We train two different 3-layer MLPs on this dataset that predict accuracy/latency respectively for a given subnet. The MLP takes the subnet's DNN architecture flattened into a vector as input. Using these predictors the search time reduces to just \textbf{2 minutes}.

\noindent \textit{Takeaway.} \figref{fig:nas_viz} shows two different DNN architectures found by \proposedfed's search stage. For AMD CPU as the target hardware, \proposedfed finds a shallow and thin DNN at 112 ms latency target. For RTX 2080Ti as the target hardware, \proposedfed finds a deep and wide DNN at 18.2 ms latency target. As RTX 2080Ti offers more floating point operations per second than AMD CPU, \proposedfed finds a more accurate specialized DNN for RTX 2080Ti (91.56\%) compared to the specialized DNN for the AMD CPU (85.25\%). \proposedfed finds specialized DNNs 
as it supports both depth and block diversity.

\begin{figure*}[tb]
	    \centering
	\begin{subfigure}[b]{0.41\textwidth}

        \includegraphics[width=1\textwidth]{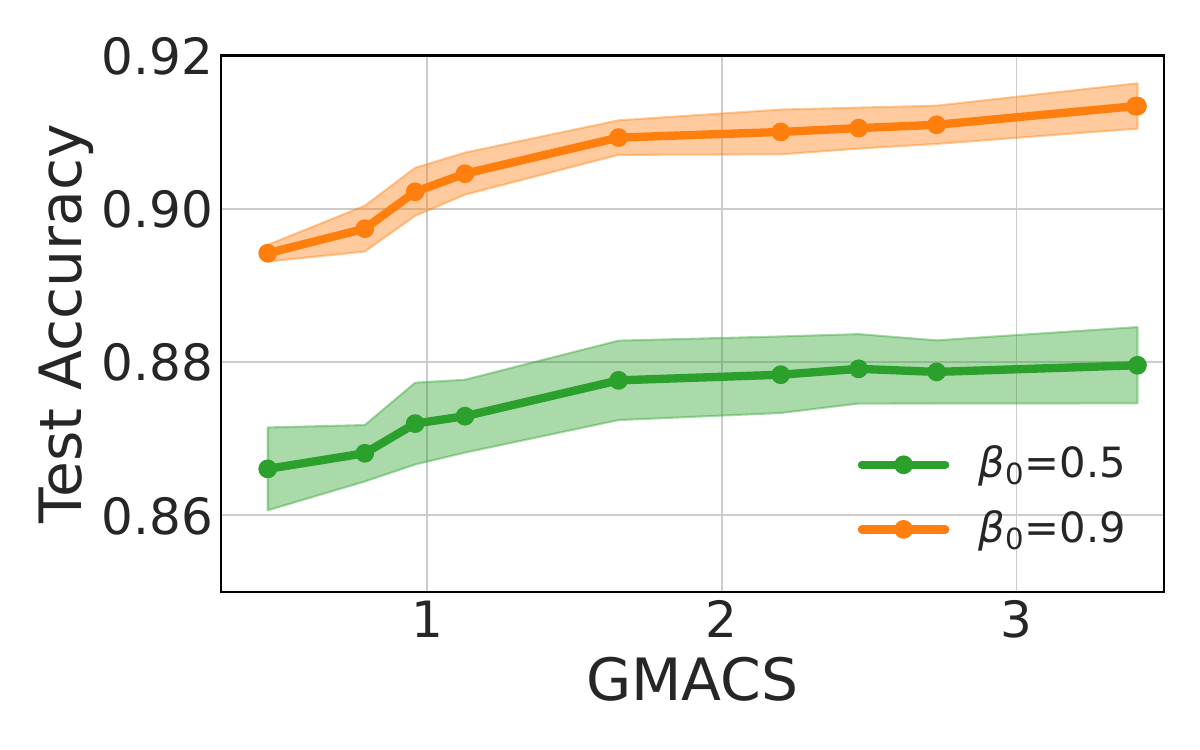}
	\vspace{-5ex}
     \caption{
		      \small  $\beta$ Init. value
		}
		\label{fig:eval:maxnet:beta}
	\end{subfigure}
	\begin{subfigure}[b]{0.41\textwidth}
        \includegraphics[width=1\textwidth]{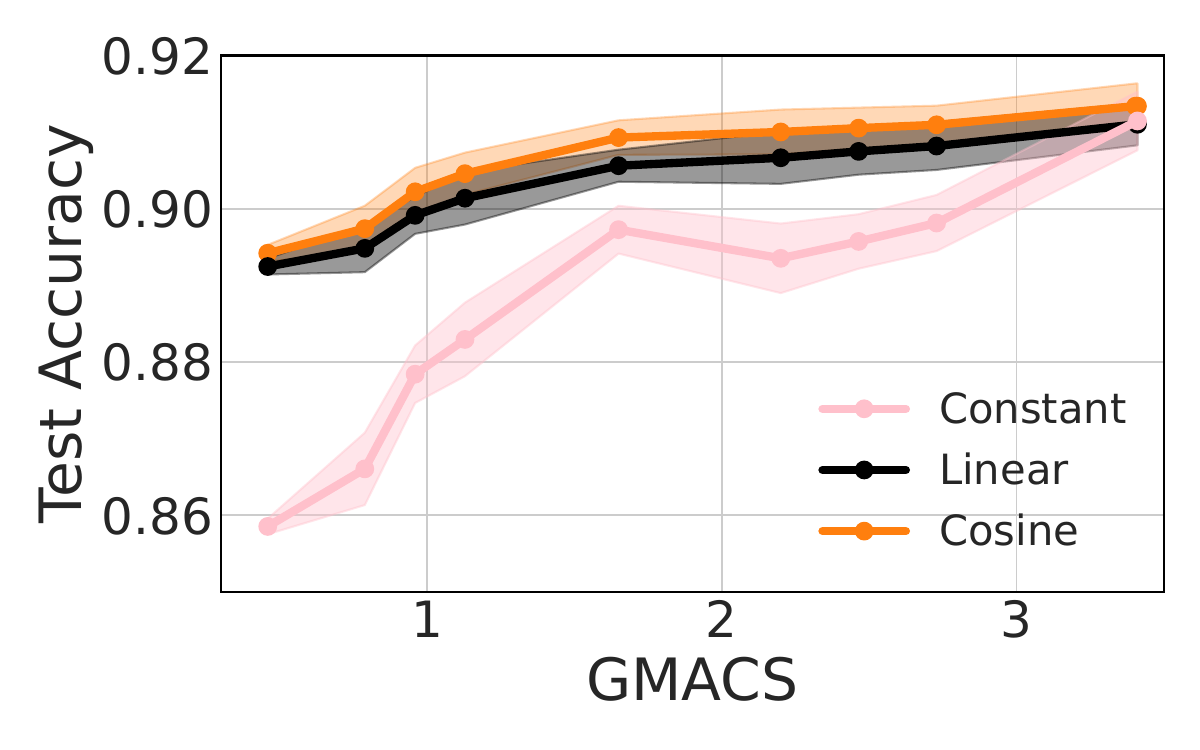}
        \vspace{-5ex}
	    \caption{
		     \small $\beta$-decay function
		}
		\label{fig:eval:maxnet:decay-fn}
	\end{subfigure}
	
	\vspace{-2ex}
	\caption{
	    \small \textbf{\spatioTempHeuristic's hyperparameters.} \textbf{a)} \proposedTraining approximates  \objref{eq:proposed_stage_fl} and performs wt. aggregation in the supernet's parameters. $\beta$ assigned to maximum subnet and $(1-\beta)$ to the rest \textbf{b)} $\beta$ is decayed from $0.9 \rightarrow$ uniform within the same number of rounds (80\% of rounds). Higher $\beta$ initially with cosine decay produces better accuracy.
	}
    \label{fig:eval:maxnet}
    \vspace{-0.2in}
\end{figure*}

\myparagraph{\proposedTraining's Hyperparameters} We provide ablation on \proposedTraining
that emulates optimization dynamics of \objref{eq:proposed_stage_fl}. 
\proposedTraining introduces two hyperparameters: the weight provided to maximum subnet's parameter in aggregation ($\beta$) and the decay function that decays $\beta$ with FL-rounds (\secref{sec:method:proposed}). \figref{fig:eval:maxnet} compares test accuracy at multiple MAC targests for different intial $\beta$ values and decay functions on CIFAR10 dataset divided into $K=20$ partitions with non-iid degree=100.

\noindent \textit{Takeaway.} The initial value of $\beta$ has a major effect on test accuracy of subnets (\figref{fig:eval:maxnet:beta}). Higher initial $\beta=0.9$ approximates the dynamics introduced via weight-shared sampling probability in \objref{eq:proposed_stage_fl} better, resulting in better accuracy. Moreover, decaying $\beta$ using the cosine function outperforms other decay functions (\figref{fig:eval:maxnet:decay-fn}). This is because the smaller subnets gradually get high loss, and therefore, get optimized in \objref{eq:proposed_stage_fl} due to $\max_{\gamma}$. This optimization dynamic is emulated by a cosine decay function over $\beta$, which gradually increases the importance of updates from smaller subnets.

\vspace{-4ex}
\section{Conclusion}
\label{sec:conclusion}
\vspace{-2ex}
\proposedfed is a scalable federated NAS method that provides efficient DNN architectures for inference deployment targets. It takes $O(1)$ training cost to satisfy $N$ deployment targets. \proposedfed achieves this by decoupling the training of DNN architectures from their search. \proposedfed's training stage uses \proposedTraining to co-train a large number of diverse DNN architectures (subnets) as part of a supernet in FL. Once the supernet is trained in FL, clients perform NAS locally with no additional training. 
\proposedTraining optimizes a novel objective that improves the performance of worst-performing subnets on each data partition.
\proposedfed is shown to surpass existing federated NAS methods. It provides optimal DNN architectures for diverse MAC targets with less training cost.

%
%
\bibliographystyle{splncs04}
\bibliography{bibs/datasets, bibs/misc, bibs/related_work, bibs/usecases}
\end{document}